\documentclass[preprint,11pt]{elsarticle}

\usepackage{longtable}

\usepackage{algorithm}
\usepackage{algpseudocode}

\usepackage{graphicx, xcolor}

\definecolor{codegray}{HTML}{f3f3f3}		
\usepackage{amssymb,amsfonts,amsmath,amsthm}
\usepackage{subcaption}
\usepackage{setspace}

\usepackage{array}
\AtBeginEnvironment{minted}{%
  }
\usepackage{geometry}
\usepackage[hidelinks,colorlinks=true,linkcolor=green,citecolor=green]{hyperref}
\usepackage{lineno}
\usepackage{booktabs,tabularx}
\theoremstyle{definition}

\newtheorem{example}{Example}[section]

\journal{arXiv}

\begin{document}

\begin{frontmatter}


\title{A Physics-Informed Machine Learning Approach for Solving Distributed Order Fractional Differential Equations}

\author{Alireza Afzal Aghaei\fnref{fn1}\corref{cor1}}
\ead{alirezaafzalaghaei@gmail.com}
\cortext[cor1]{Corresponding author}
\fntext[fn1]{Independent researcher, Isfahan, Iran}



\begin{abstract}
This paper introduces a novel methodology for solving distributed-order fractional differential equations using a physics-informed machine learning framework. The core of this approach involves extending the support vector regression (SVR) algorithm to approximate the unknown solutions of the governing equations during the training phase. By embedding the distributed-order functional equation into the SVR framework, we incorporate physical laws directly into the learning process. To further enhance computational efficiency, Gegenbauer orthogonal polynomials are employed as the kernel function, capitalizing on their fractional differentiation properties to streamline the problem formulation. Finally, the resulting optimization problem of SVR is addressed either as a quadratic programming problem or as a positive definite system in its dual form. The effectiveness of the proposed approach is validated through a series of numerical experiments on Caputo-based distributed-order fractional differential equations, encompassing both ordinary and partial derivatives.
\end{abstract}


\end{frontmatter}


\section{Introduction}
Machine learning, a critical branch of artificial intelligence (AI), is transforming modern research and industry by enabling data-driven decision-making and predictive analytics. Central to machine learning is regression analysis, a fundamental tool that models the relationship between variables \cite{rad2023learning}. This technique is essential not only for making predictions but also for uncovering underlying patterns within data. Through methods ranging from simple linear regression to more sophisticated approaches like regularization and kernel-based techniques, regression provides the foundation for both straightforward and complex data modeling tasks.

Physics-informed machine learning (PIML) is an emerging field that integrates physical laws and principles into machine learning frameworks to enhance model accuracy and reliability, particularly in scientific and engineering applications \cite{babaei2024solving}. In PIML, regression plays a key role in approximating the solutions to both forward and inverse problems. Forward problems involve predicting system behavior based on known inputs, while inverse problems aim to infer unknown inputs from observed outputs \cite{aghaei2024pinnies}. Regression techniques are employed to approximate the solution space, ensuring that the model adheres to known physical laws, such as the conservation of energy or mass. By embedding these constraints, PIML models can achieve higher accuracy and robustness, making them especially valuable in scenarios where data is sparse or noisy, and where traditional numerical methods might struggle. This approach is particularly useful in fields like fluid dynamics, material science, and climate modeling, where the interplay between data-driven insights and physical principles is crucial for making reliable predictions \cite{abbaszadeh2024supervised,aghaei2024pinnies}.

In tackling physics-informed machine learning tasks, various machine learning regression techniques, such as Extreme Learning Machines (ELM), Support Vector Regression (SVR), and neural networks, are employed to model complex systems \cite{rad2023learning}. ELMs are known for their speed and efficiency in training, as they utilize a random selection of hidden nodes and require no iterative tuning. Neural networks, with their ability to learn complex, non-linear relationships, are powerful but often require significant computational resources and large datasets to achieve high accuracy. However, among these techniques, SVR stands out due to its unique property of maximizing the margin of the data, which leads to better generalization and higher accuracy in solving problems, particularly in PIML tasks. SVR's ability to maintain high precision even with small datasets makes it highly effective for regression tasks in PIML, where data might be limited or noisy \cite{an2010study, babaei2024solving}.

SVR has demonstrated exceptional accuracy in solving regression tasks, prompting the development of various extensions aimed at enhancing its performance and applicability. Notable among these are Twin Support Vector Machines, Fuzzy Support Vector Machines, and Least Squares Support Vector Regression (LSSVR). The LSSVR algorithm, in particular, replaces the conventional loss function with a squared loss, simplifying the mathematical formulation and significantly speeding up the training process. This efficiency makes LSSVR especially suitable for large-scale and complex problems where computational resources are a concern. Recently, LSSVR has gained considerable attention for its ability to solve forward physics-informed mathematical problems, including ordinary and partial differential equations, integral equations, fractional differential and integro-differential equations, delay problems, and differential algebraic equations. The timeline in \ref{tbl:literature-review} provides a comprehensive overview of the evolution and application of the LSSVR method across these domains, illustrating its growing prominence in the field.

In recent years, as reflected in the table, the focus of research has increasingly shifted toward fractional problems, highlighting the growing importance of fractional calculus—a branch of calculus that extends the concept of derivatives and integrals to non-integer orders. Fractional differential operators, such as the Caputo, Riemann-Liouville, and Grünwald-Letnikov derivatives, are at the forefront of this research due to their ability to model memory and hereditary properties inherent in many physical systems \cite{nosrati2024machine, abbas2014fractional, mainardi2022fractional}. Among these, the Caputo derivative is particularly notable for its application in initial value problems, as it allows for the inclusion of traditional boundary conditions, making it more suitable for real-world applications \cite{nosrati2024machine}.

A review of the literature reveals that Caputo-based fractional differential equations (FDEs) have been extensively studied for their use in modeling viscoelastic materials, anomalous diffusion, and complex dynamic systems \cite{aghaei2024pinnies}. Variable order and distributed order fractional differential equations are two important generalizations of fractional differential equations that have gained increasing attention in recent years \cite{moghaddam2019numerical,morgado2015numerical, katsikadelis2014numerical}. Variable order fractional differential equations (VOFDEs) involve fractional derivatives or integrals whose orders are functions of time, space, or other variables, rather than constants. Meanwhile, distributed order fractional differential equations (DOFDEs) involve an integration of fractional derivatives over a range of orders. These generalizations allow for more flexible modeling of complex systems with memory effects that vary over time or involve multiple scales. VOFDEs can capture phenomena where the memory effect changes with time or other variables, while DOFDEs can model systems with a continuous spectrum of time scales or memory effects \cite{katsikadelis2014numerical,moghaddam2019numerical}.

As the application of FDEs has expanded, solving these complex problems has garnered significant attention. Given that analytical solutions are often intractable, researchers have developed various efficient numerical methods to address these challenges. These include spectral methods, meshless methods, and finite difference schemes, each tailored to efficiently and accurately solve advanced FDEs \cite{moghaddam2019numerical,morgado2015numerical, katsikadelis2014numerical,abbaszadeh2021meshless}. Meshless methods, for instance, utilize radial basis functions to approximate solutions without the need for a predefined mesh, making them particularly flexible for complex geometries \cite{abbaszadeh2021meshless}. Spectral methods, on the other hand, rely on orthogonal polynomials such as Chebyshev polynomials, Legendre polynomials, and Jacobi polynomials, which are well-suited for problems with smooth solutions and often lead to more accurate results due to their exponential convergence properties \cite{rashidinia2021novel}. In the context of machine learning, these orthogonal polynomials serve as basis functions, analogous to the concept of a feature map, which is used to transform input data into a higher-dimensional space. This transformation is integral to the function of kernel methods in machine learning, particularly in SVR. The kernel function, central to SVR, computes the inner product of these basis functions (or feature maps) in the transformed space, enabling the formulation of the regression problem in a way that captures complex, nonlinear relationships. This approach not only facilitates the handling of high-dimensional data but also enhances the accuracy and generalization capability of the model, making kernel functions a powerful tool in both numerical analysis of FDEs and machine learning applications \cite{rad2023learning}.

In this paper, we propose the use of Gegenbauer polynomials as the kernel function within the physics-informed machine learning form of the LSSVR framework for solving forward forms of distributed-order fractional differential equations. Gegenbauer polynomials generalize Legendre and Chebyshev polynomials and possess unique properties that make them well-suited for approximating solutions to FDEs. By leveraging the fractional differentiation properties of these polynomials, our method simplifies the problem formulation and improves computational efficiency, offering a robust and accurate solution to the complexities associated with distributed-order fractional differential equations. Specifically, our contribution is as follows:
\begin{enumerate}
    \item Deriving an LS-SVR method for distributed-order fractional differential equations.
    \item Employing Gegenbauer polynomials as the kernel function in LSSVR.
    \item Simulating one-dimensional DOFDEs using the proposed approach.
    \item Solving DOFDEs with partial derivatives using the developed framework.
    \item Performing hyperparameter tuning and sensitivity analysis on the Gegenbauer parameter during the numerical solution process.
    
\end{enumerate}
The remainder of the article is structured as follows: Section 2 covers the prerequisites related to this study. In Section 3, we derive the LS-SVR approach for DOFDEs. Section 4 presents examples, numerical results, comparison tables, and figures related to these results. Finally, in Section 5, we discuss the significant impact of the proposed method on achieving high accuracy in solving these problems.
 \renewcommand{\arraystretch}{0.8} 

\begin{longtable}{@{}llll p{4.5cm}@{}}
\caption{A timeline on LS-SVR method for solving various types of functional equations such as Fractional, Partial Differential Equations (PDE), Ordinary Differential Equations (ODE), Systems of Ordinary Differential Equations (Sys. ODE), Systems of Integral Equations (Sys. IE), Integral Equations (IE), Volterra Integral Equations (VIE), Volterra-Fredholm Integral Equations (VFIE), Fredholm Integral Equations (FIE), Inverse Partial Differential Equations (Inv. PDE), Stochastic Differential Equations (SDE), Volterra Integro-Differential Equations (VIDE), Fractional Integro-Differential Equations (FIDE), Delay Differential Equations (DDE), Differential-Algebraic Equations (DAE), Fractional Differential-Algebraic Equations (Frac. DAE), Fractional Ordinary Differential Equations (Frac ODE), and Systems of Fractional Differential Equations (Sys. FDE).}
\label{tbl:literature-review}
\\
\toprule
Authors & Year & Problem type & Domain & Kernel \\ \midrule
\endfirsthead

\multicolumn{5}{c}%
{{\bfseries Continued from previous page}} \\ \toprule
Authors & Year & Problem type & Domain & Kernel \\ \midrule
\endhead

\midrule
\multicolumn{5}{r}{{Continued on next page}} \\
\endfoot

\bottomrule
\endlastfoot
\citeauthor{an2010study} & 2010 & PDE & Finite & Standard Polynomial \\
\citeauthor{krebs2011support} & 2011 & FIE & Finite & RBF \\
\citeauthor{mehrkanoon2012approximate} & 2012 & ODE & Finite & RBF \\
\citeauthor{mehrkanoon2012parameter} & 2012 & ODE & Finite & RBF \\
\citeauthor{guo2012ls} & 2012 & VIE & Finite & RBF \\
\citeauthor{mehrkanoon2012ls} & 2012 & DAE & Finite & RBF \\
\citeauthor{zhang2013new} & 2013 & ODE & Finite & RBF \\
\citeauthor{mehrkanoon2013ls} & 2013 & DDE & Finite & RBF \\
\citeauthor{zhang2014ls} & 2014 & Sys. ODE & Finite & RBF \\
\citeauthor{mehrkanoon2015learning} & 2015 & PDE & Finite & RBF \\
\citeauthor{wu2016approximate} & 2016 & PDE & Finite & PDE \\
\citeauthor{wu2017approximate} & 2017 & PDE & Finite & RBF \\
\citeauthor{dong2018multilevel} & 2018 & PDE & Finite & Wavelet \\
\citeauthor{yu2018approximate} & 2018 & PDE & Finite & RBF \\
\citeauthor{han2018learning} & 2018 & PDE & Finite & RBF \\
\citeauthor{leake2019analytically} & 2019 & ODE/PDE & Finite & RBF \\
\citeauthor{han2019learning} & 2019 & PDE & Finite & Finite-Elements/RBF \\
\citeauthor{dong2019wavelet} & 2019 & PDE & Finite & Wavelet \\
\citeauthor{lu2019ls} & 2019 & ODE & Finite & RBF \\
\citeauthor{wu2019learning} & 2019 & Inv. PDE & Finite & RBF \\
\citeauthor{lu2020solving} & 2020 & ODE & Finite & RBF \\
\citeauthor{hajimohammadi2020new} & 2020 & ODE & Semi-Infinite & Rational Gegenbauer \\
\citeauthor{parand2021new} & 2021 & FIE & Finite & Legendre \\
\citeauthor{parand2021parallel} & 2021 & Frac. VIDE& Semi-Infinite & Fractional Rational Legendre \\
\citeauthor{pakniyat2021least} & 2021 & ODE & Infinite & Hermite \\
\citeauthor{parand2021numerical} & 2021 & VFIE & Finite & Legendre \\
\citeauthor{wu2021learning} & 2021 & PDE & Bounded & RBF \\
\citeauthor{hajimohammadi2021numerical} & 2021 & Frac. PDE& Semi-infinite & Laguerre \\
\citeauthor{parand2022machine} & 2022 & Inv. PDE & Finite & Chebyshev \\
\citeauthor{pakniyat2022numerical} & 2022 & ODE & Semi-infinite & Hermite \\
\citeauthor{khoee2022least} & 2022 & ODE & Finite & Legendre \\
\citeauthor{rahimkhani2022chelyshkov} & 2022 & SDE & Finite & Wavelet \\
\citeauthor{ahadian2022support} & 2022 & Frac. PDE& Finite & Bernstein \\
\citeauthor{parand2022least} & 2022 & VIE & Finite & Legendre \\
\citeauthor{hadian2023solving} & 2023 & DOFDE & Finite & Legendre \\
\citeauthor{moayeri2023solving} & 2023 & PDE & Finite & Legendre \\
\citeauthor{hajimohammadi2023novel} & 2023 & ODE & Semi-infinite & Laguerre \\
\citeauthor{taheri2023bridging} & 2023 & DDE& Finite & Legendre \\
\citeauthor{aghaei2023hyperparameter} & 2023 & ODE & Semi-infinite & Fractional Rational Jacobi \\
\citeauthor{mehrdad2023numerical} & 2023 & Sys. ODE & Finite & RBF \\
\citeauthor{parand2023solving} & 2023 & Sys. IE & Finite & Legendre \\
\citeauthor{razzaghi2023solving} & 2023 & ODE & Semi-infinite & Rational Legendre \\
\citeauthor{shivanian2023novel} & 2023 & ODE & Finite & Gegenbauer \\
\citeauthor{rahimkhani2023performance} & 2023 & ODE & Finite & Genocchi wavelet \\
\citeauthor{rahimkhani2023bernoulli} & 2023 & Sys. FDE & Finite & Wavelet \\
\citeauthor{sun2024novel} & 2024 & FIDE & Finite & Standard polynomial \\
\citeauthor{abbaszadeh2024supervised} & 2024 & Frac. PDE& Finite & RBF \\
\citeauthor{sun2024numerical} & 2024 & VFIE & Finite & Standard Polynomial \\ 
\citeauthor{babaei2024solving} & 2024 & ODE & Semi-Infinite & Fractional Rational Chebyshev \\ 
\citeauthor{mohammadi2024new} & 2024 & PDE & Finite & Bernstein \\
\citeauthor{mohammadi2024space} & 2024 & Inv. PDE & Finite & Bernstein \\
\citeauthor{ordokhani2024application} & 2024 & Frac ODE & Finite & Wavelet \\
\citeauthor{nosrati2024machine} & 2024 & Frac. PDE & Finite & Legendre \\
\citeauthor{taheri2024new} & 2024 & Frac. DAE & Finite & Legendre \\
\bottomrule
\end{longtable}
\renewcommand{\arraystretch}{1}
\section{Background}
In this section, we provide the necessary mathematical background for the subsequent sections, where we present our approach.

\subsection{Gegenbauer polynomials}
Gegenbauer polynomials, also known as ultraspherical polynomials, are a class of orthogonal polynomials that generalize the Legendre and Chebyshev polynomials. They are widely used in mathematical physics, particularly in the study of spherical harmonics and solutions to the Laplace equation in higher dimensions \cite{rad2023learning}. Gegenbauer polynomials have significant applications in mathematical physics, especially in problems involving spherical symmetry \cite{rad2023learning,hajimohammadi2020new,shivanian2023novel}. For example, they appear in the expansion of Green's function of the Laplace equation in spherical coordinates, as well as in the solution of the Helmholtz equation. The Gegenbauer polynomials \( C_n^{(\lambda)}(t) \) are expressed in terms of the hypergeometric function as:
\[
C_n^{(\lambda)}(t) = \frac{(2\lambda)_n}{n!} \, {}_2F_1\left(-n, 2\lambda + n; \lambda + \frac{1}{2}; \frac{1-t}{2}\right),
\]
where \((a)_n\) denotes the Pochhammer symbol, representing the rising factorial. They can also be generated using a three-term recurrence relation:
\begin{equation*}
    \begin{aligned}
    &(n+1) C_{n+1}^{(\lambda)}(t) = 2t(n+\lambda) C_n^{(\lambda)}(t) - (n+2\lambda-1) C_{n-1}^{(\lambda)}(t),\\
&C_0^{(\lambda)}(t) = 1, \quad C_1^{(\lambda)}(t) = 2\lambda t,
\end{aligned}
\end{equation*}
or an explicit formula:
\begin{equation}
    {\displaystyle C_{n}^{(\lambda )}(t)=\sum _{k=0}^{\lfloor n/2\rfloor }(-1)^{k}{\frac {\Gamma (n-k+\lambda )}{\Gamma (\lambda )k!(n-2k)!}}(2t)^{n-2k}.}
    \label{eq:explicit}
\end{equation}
Using this formula, the first few Gegenbauer polynomials can be obtained as:
\begin{align*}
C_0^{(\lambda)}(t) &= 1, \\
C_1^{(\lambda)}(t) &= 2\lambda t, \\
C_2^{(\lambda)}(t) &= \left(2 \lambda t^{2}+2 t^{2}-1\right) \lambda , \\
C_3^{(\lambda)}(t) &= \frac{1}{3} 4 t \lambda \left(\lambda t^{2}+2 t^{2}-\frac{3}{2}\right) \left(\lambda +1\right), \\
C_4^{(\lambda)}(t) &= \frac{1}{3} 2 \lambda \left(\frac{3}{4}+\left(\lambda^{2}+5 \lambda +6\right) t^{4}+\left(-3 \lambda -6\right) t^{2}\right) \left(\lambda +1\right).
\end{align*}

Gegenbauer polynomials are orthogonal with respect to the weight function \((1-t^2)^{\lambda-\frac{1}{2}}\) on the interval \([-1, 1]\) for \(\lambda > -\frac{1}{2}\). Specifically, they satisfy the following orthogonality relation:
\[
\int_{-1}^{1} (1-t^2)^{\lambda-\frac{1}{2}} C_m^{(\lambda)}(t) C_n^{(\lambda)}(t) \, dt = 
\begin{cases} 
0 & \text{if } m \neq n, \\
\frac{\pi \, 2^{1-2\lambda} \Gamma(n+2\lambda)}{n!(n+\lambda)\Gamma(\lambda)^2} & \text{if } m = n,
\end{cases}
\]
where $\Gamma(\lambda)$ is the Gamma function given by \(\Gamma(z) = \int_0^\infty t^{z-1} e^{-t} \, dt\). This orthogonality can be shifted to any finite domain \([a,b]\) by applying an affine transformation of formula \(\mu(t) = \frac{2t-a-b}{b-a}\) to the input of Gegenbauer polynomials, i.e. \(G^{(\lambda)}(t) = C^{(\lambda)}(\mu(t))\).

The derivatives of Gegenbauer polynomials with respect to \(t\) are given by:
\begin{equation}
    \frac{d}{dt} C_n^{(\lambda)}(t) = 2\lambda C_{n-1}^{(\lambda+1)}(t).
    \label{eq:gegdiff}
\end{equation}
This relation can be used to derive further properties of the polynomials, particularly in applications involving ordinal, partial, and fractional differential equations. Additionally, the derivatives with respect to the parameter \(\lambda\) are also of interest:
\[
\frac{\partial}{\partial \lambda} C_n^{(\lambda)}(t) = \sum_{k=0}^{n-1} C_k^{(\lambda)}(t) \frac{1}{\lambda+k}.
\]

\subsection{Fractional Calculus}

Fractional calculus is a generalization of classical calculus to non-integer orders of differentiation and integration. It extends the concept of derivatives and integrals to arbitrary (real or complex) orders, providing powerful tools for modeling processes with memory and hereditary properties. In the following re recall some of the well-known definitions and formulas related to them, which will be used in the methodology.

The Riemann-Liouville integral is a fundamental concept in fractional calculus, defined for a function \( u(t) \) and a real number \(\alpha > 0\) as follows:
\[
I_t^\alpha u(t) = \frac{1}{\Gamma(\alpha)} \int_0^t (t - \tau)^{\alpha - 1} u(\tau) \, d\tau,
\]
where \( \Gamma(\alpha) \) denotes the Gamma function. Similar to the integral operator, the Riemann-Liouville fractional derivative for \(\alpha > 0\) is defined as:
\[
^{RL}D_t^\alpha u(t) = \frac{1}{\Gamma(n - \alpha)} \frac{d^n}{dt^n} \int_0^t (t - \tau)^{n - \alpha - 1} u(\tau) \, d\tau,
\]
where \( n = \lceil \alpha \rceil \) is the smallest integer greater than or equal to \( \alpha \). This derivative generalizes the classical derivative to non-integer orders \cite{nosrati2024machine}. However, this definition introduces complexities when modeling the initial values of differential equations. Therefore, we use the Caputo derivative definition, which facilitates a more straightforward interpretation of initial value problems. This derivative is defined as:
\[
^C D_t^\alpha u(t) = \frac{d^\alpha}{dt^\alpha}u(t) = \frac{1}{\Gamma(n - \alpha)} \int_0^t (t - \tau)^{n - \alpha - 1} \frac{d^n}{d\tau^n} u(\tau) \, d\tau,
\]
where \( n = \lceil \alpha \rceil \). The Caputo fractional derivative has several important properties:
\begin{itemize}
    \item \textbf{Linearity:} For functions \( u(t) \) and \( v(t) \), and constants \( a \) and \( b \),
    \[
    ^C D_t^\alpha \left[ a u(t) + b v(t) \right] = a \, ^C D_t^\alpha u(t) + b \, ^C D_t^\alpha v(t).
    \]

    \item \textbf{Initial Conditions:} The Caputo derivative of a constant is zero:
    \[
    ^C D_t^\alpha c = 0 \text{ for } c \in \mathbb{R}.
    \]
    
    \item \textbf{Derivative of Polynomials:} The Caputo derivative of polynomial \( u(t) = t^n \) is given by:
    \begin{equation}
        ^C D_t^\alpha t^n = \frac{\Gamma(n+1)}{\Gamma(n-\alpha+1)} t^{n-\alpha}.
        \label{eq:capmon}
    \end{equation}
    
    \item \textbf{Special Cases:} When \(\alpha\) is an integer, the Caputo derivative reduces to the classical derivative:
    \[
    ^C D_t^n u(t) = \frac{d^n u(t)}{dt^n}.
    \]
\end{itemize}

\section{The Proposed Approach}
In this section, we consider the following distributed-order fractional differential equation:
\begin{equation}
\psi\left[t, u(t), ^{C}D^{\eta} u(t)\right] = \rho(t) + \int_{a}^{b} \phi(\theta) \frac{d^\theta}{dt^\theta} u(t) \, d\theta,
\label{eq:distrib}
\end{equation}
where $\psi(\cdot)$, $\phi(\cdot)$, and $\rho(\cdot)$ are known functions, $a, b \in \mathbb{R}$ are the bounds of the integration, and $u(t)$ is the unknown function. To approximate the solution to this problem, we consider a linear combination of some unknown weights and Gegenbauer polynomials:
\begin{equation}
    \hat{u}(t) = \sum_{i=0}^{d-1} \mathbf{w}_i G_i^{(\lambda)}(t),
    \label{eq:approx}
\end{equation}
where $d$ is the number of basis functions, $w_i$ are the unknown weights, and $C_i^{(\lambda)}(t)$ are the Gegenbauer polynomials. In the case of a partial DOFDE with two independent variables, this expansion takes the form:
\begin{equation*}
    \hat{u}(x,t) = \sum_{i=0}^{d_x-1} \sum_{j=0}^{d_t-1} \mathbf{w}_{i,j} G_i^{(\lambda)}(x) G_j^{(\lambda)}(t).
\end{equation*}
However, by vectorization of matrix $\mathbf{w}\in \mathbb{R}^{d_x-1\times d_t-1}$ arrangement of shifted Gegenbauer polynomials, one can rewrite the two-dimensional approximation in the form of Equation \eqref{eq:approx}. In either case, we formulate the following optimization problem using the Least-Squares Support Vector Regression framework \cite{aghaei2023hyperparameter, parand2021parallel}:
\begin{align*}
    \min_{w,e} \quad & \frac{1}{2} \mathbf{w}^T \mathbf{w} + \frac{\gamma}{2} \mathbf{e}^T \mathbf{e} \\
    \text{subject to} \quad & \psi\left[t_i, \hat{u}(t_i), \frac{d^\eta}{dt^\eta} \hat{u}(t_i)\right] - \rho(t_i) \\
    & \quad - \int_{a}^{b} \phi(\theta) \frac{d^\theta}{dt^\theta} \hat{u}(t_i) \, d\theta = \mathbf{e}_i, \quad i = 1, \dots, N,
\end{align*}
where $N$ is the number of training points, $\gamma$ is a regularization parameter, and $\mathbf{e}_i$ represents the residual error terms.

In numerical simulations, computing the analytical integration can be challenging. To mitigate this issue, we first approximate the integral using the accurate Gauss-Legendre quadrature of order $Q$, which converts the integral into a finite summation:
\begin{equation*}
    \int_{a}^{b} \phi(\theta) \frac{d^\theta}{dt^\theta} \hat{u}(t_i) \, d\theta \approx \frac{b-a}{2} \sum_{j=0}^{Q} \omega_j \phi(\hat{\theta}_j) \frac{d^{\hat{\theta}_j}}{dt^{\hat{\theta}_j}} \hat{u}(t_i),
\end{equation*}
where the nodes $\hat{\theta}_j$ are given by:
\[
\hat{\theta}_j = \frac{b-a}{2} \theta_j + \frac{a+b}{2},
\]
in which $\theta_j$ are the roots of Legendre polynomial $G^{(0)}$ and $\omega_j$ are the Gauss-Legendre weights corresponding to the nodes $\theta_j$ given by:
\begin{equation*}
    \omega_j ={\frac {2}{\left(1-x_{j}^{2}\right)\left[{G^{(0)\prime}}_{Q}(x_{j})\right]^{2}}}.
\end{equation*}
Employing this technique helps us reduce the computational complexity of evaluating an analytical integral to a finite summation. Moreover, this method allows us to utilize the Caputo fractional derivative property of polynomials, as given in Equation \eqref{eq:capmon}, along with the linearity of the approximation:
\begin{equation*}
\begin{aligned}
        \frac{d^{\hat{\theta}_j}}{dt^{\hat{\theta}_j}} \hat{u}(t) &= \sum_{i=0}^{d-1} w_i \frac{d^{\hat{\theta}_j}}{dt^{\hat{\theta}_j}} C_i^{(\lambda)}(t)\\
        &=\sum_{i=0}^{d-1} w_i \frac{d^{\hat{\theta}_j}}{dt^{\hat{\theta}_j}} \Bigg[{\displaystyle\sum _{k=0}^{\lfloor i/2\rfloor }(-1)^{k}{\frac {\Gamma (i-k+\lambda )}{\Gamma (\lambda )k!(i-2k)!}}(2t)^{i-2k}}\Bigg] \\
        &= \sum_{i=0}^{d-1} w_i {\displaystyle\sum _{k=0}^{\lfloor i/2\rfloor }(-1)^{k}{\frac {\Gamma (i-k+\lambda )}{\Gamma (\lambda )k!(i-2k)!}}\frac{d^{\hat{\theta}_j}}{dt^{\hat{\theta}_j}}\Big[(2t)^{i-2k}\Big]} \\
        &= \sum_{i=0}^{d-1} w_i {\displaystyle\sum _{k=0}^{\lfloor i/2\rfloor }(-1)^{k}2^{i-2k} {\frac {\Gamma (i-k+\lambda )}{\Gamma (\lambda )k!(i-2k)!}}\frac{\Gamma(i-2k+1)}{\Gamma(i-2k-\alpha+1)} t^{i-2k-\alpha}},
\end{aligned}
\end{equation*}
which facilitates fast computation of the derivatives of the unknown function. Combining all these, the optimization problem can be reformulated as:
\begin{equation*}
    \min_{w} \, \frac{1}{2} \boldsymbol{w}^T \boldsymbol{w} + \frac{\gamma}{2} \boldsymbol{e}^T \boldsymbol{e},
\end{equation*}
subject to:
\begin{equation*}
    \psi\left[t_i, \hat{u}(t_i), \frac{d^\eta}{dt^\eta} \hat{u}(t_i)\right] - \rho(t_i) - \frac{b-a}{2} \sum_{j=0}^{I} \omega_j \phi(\hat{\theta}_j) \frac{d^{\hat{\theta}_j}}{dt^{\hat{\theta}_j}} \hat{u}(t_i) = e_i, \quad i = 1, \ldots, N,
\end{equation*}
for $N$ training points. This quadratic optimization problem can be reformulated in the dual space. To achieve this, we first construct the Lagrangian function:
\begin{equation*}
    \begin{aligned}
        \mathfrak{L}(\boldsymbol{w}, \boldsymbol{e}, \boldsymbol{\beta}) =& \frac{1}{2} \boldsymbol{w}^T \boldsymbol{w} + \frac{\gamma}{2} \boldsymbol{e}^T \boldsymbol{e}\\ &+ \sum_{i=1}^{N} \beta_i \left(\psi\left[t_i, \hat{u}(t_i), \frac{d^\eta}{dt^\eta} \hat{u}(t_i)\right] - \rho(t_i) - \frac{b-a}{2} \sum_{j=0}^{I} \omega_j \phi(\hat{\theta}_j) \frac{d^{\hat{\theta}_j}}{dt^{\hat{\theta}_j}} \hat{u}(t_i) - e_i\right),
    \end{aligned}
\end{equation*}
where $\beta_i$ are the Lagrange multipliers associated with the constraints. Next, we derive the Karush-Kuhn-Tucker (K.K.T.) conditions, which yield:
\begin{equation*}
    \Bigg[\boldsymbol{Z}^T \boldsymbol{Z} + \frac{1}{\gamma} \boldsymbol{I}\Bigg] \boldsymbol{\beta}  = \boldsymbol{\rho},
\end{equation*}
where $\boldsymbol{\rho}_i = \rho(t_i)$ and the elements of matrix $\boldsymbol{Z}$ is defined as:
\begin{equation*}
    Z_{i,j} = \psi\left[t_i, \hat{u}(t_i), \frac{d^\eta}{dt^\eta} \hat{u}(t_i)\right] - \rho(t_i) - \frac{b-a}{2} \sum_{k=0}^{I} \omega_k \phi(\hat{\theta}_k) \frac{d^{\hat{\theta}_k}}{dt^{\hat{\theta}_k}} C_j^{(\lambda)}(t_i),
\end{equation*}
with the quadrature points $\hat{\theta}_k$ and weights $\omega_k$ determined by the Gauss-Legendre quadrature method. By solving this positive definite system of equations, the unknown In the dual space, the Lagrangian multipliers $\boldsymbol{\beta}$ are fixed and then the approximation can be expressed as $\hat{u} = \boldsymbol{\beta}^T\boldsymbol{Z}^T \mathbf{G}(t)$ which gives the solution in terms of Gegenbauer kernel function:
\begin{equation*}
    \hat{u}(t) = \sum_{i=1}^N \beta_i \mathcal{L}K(t, t_i),
\end{equation*}
where $\mathcal{L}$ is the given problem in operator form and $K(t,t_i)$ is defined as
\begin{equation*}
    K(t,t_i) = \sum_{j=0}^d G^{(\lambda)}_j(t) G^{(\lambda)}_j(t_i).
\end{equation*}
\section{Numerical examples}
In this section, we simulate some DOFDEs using the proposed approach. The problems are chosen to ensure their analytical solutions cover various function spaces, including polynomials, fractional functions, and one example with no known exact solution. All experiments are implemented using Maple Mathematical Software and run on a personal computer with an Intel Core i3-10100F processor and 16 GB of RAM.

\subsection{Ordinal DOFDEs}
In this section, we examine two DOFDEs in one-dimensional space. For all of the problems in this section, we consider Gaussian quadrature discretization $Q=10$.
\begin{example}
\label{ex:1}
The following distributed differential equation problem, with the analytical solution \( u(t) = t^2 \) and initial conditions \( u(0) = u'(0) = 0 \), is discussed in \cite{katsikadelis2014numerical}. The problem is given by:
\begin{equation*}
\int_{0.2}^{1.5} \Gamma(3-\theta) \, ^CD^\theta \, u(t) \, d\theta = 2 \frac{t^{1.8} - t^{0.5}}{\ln t},
\end{equation*}
where \( D^\theta \) denotes the distributed-order fractional derivative of \( u(t) \) with respect to \( t \), and \( \Gamma \) represents the Gamma function. We solve this problem using the proposed LSSVR approach in the domain \( t \in [0.2,1.5] \) with \( d=4 \), and \( N=4 \). To determine the optimal value for $\gamma$, we conducted a hyperparameter analysis following the approach outlined in \cite{aghaei2023hyperparameter}. A random search algorithm was employed to optimize $\lambda$, aiming to minimize the residual error. The results of this sensitivity analysis are presented in Figure \ref{fig:hpspace}, which indicates that $\lambda$ has minimal influence on the solution. Therefore, we select a simplified value, such as $0$ or $\frac{1}{2}$, for ease of formulation. The simulation results with this hyperparameter choice are displayed in Figure \ref{fig:ex1}, demonstrating excellent accuracy, particularly because the exact solution is a polynomial, consistent with the basis functions used.
\begin{figure}
    \centering
    \includegraphics[width=0.8\linewidth]{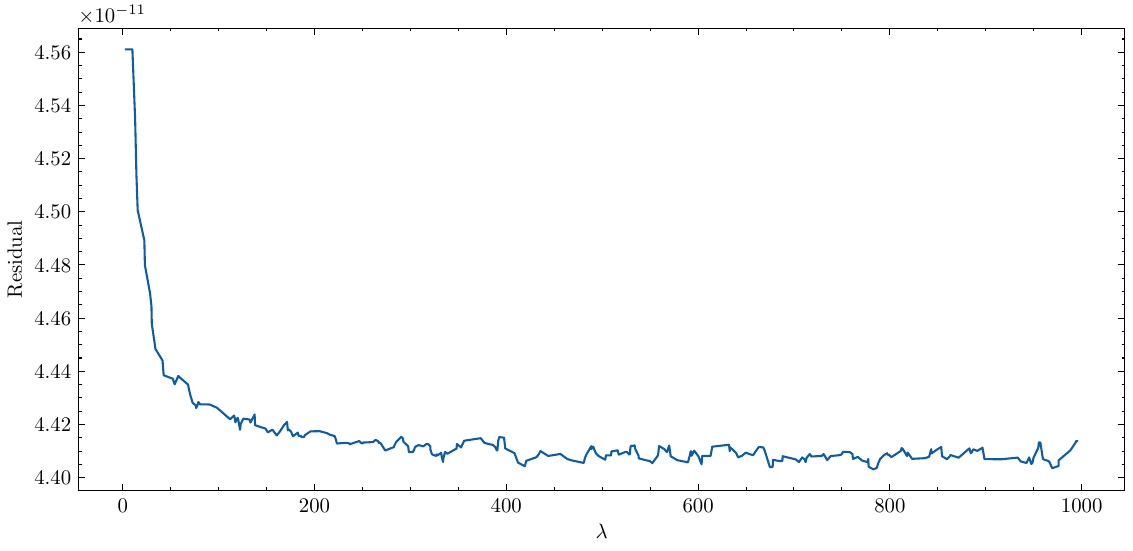}
    \caption{Hyperparameter space for $\lambda$ in Gegenbauer polynomials $G^{(\lambda)}(t)$ used for solving Example \ref{ex:1}.}
    \label{fig:hpspace}
\end{figure}
\begin{figure}[ht]
    \centering
    \begin{subfigure}[b]{0.45\textwidth}
        \centering
        \includegraphics[width=\textwidth]{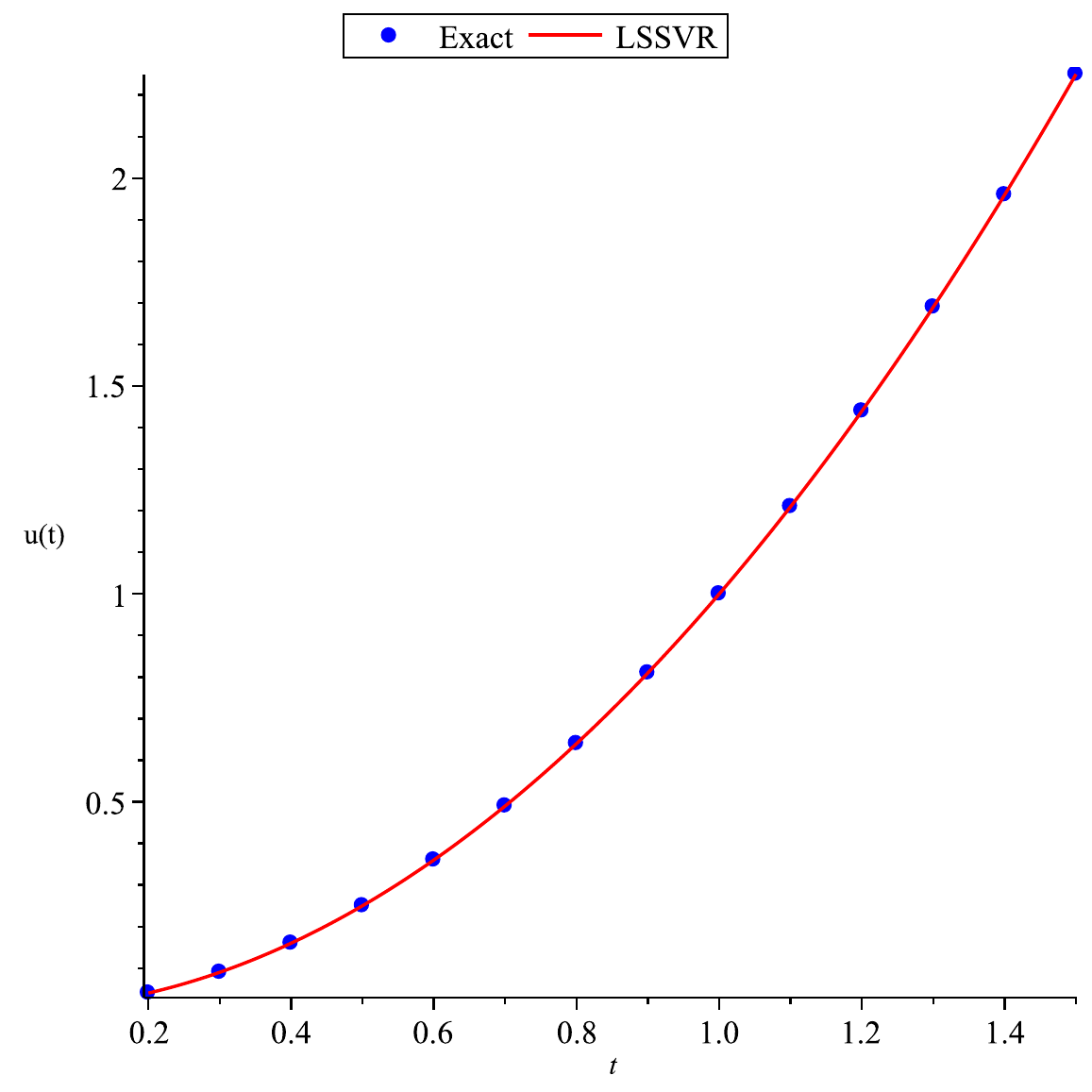}
   \caption{Predicted solution}
    \end{subfigure}
    \hfill
    \begin{subfigure}[b]{0.45\textwidth}
        \centering
        \includegraphics[width=\textwidth]{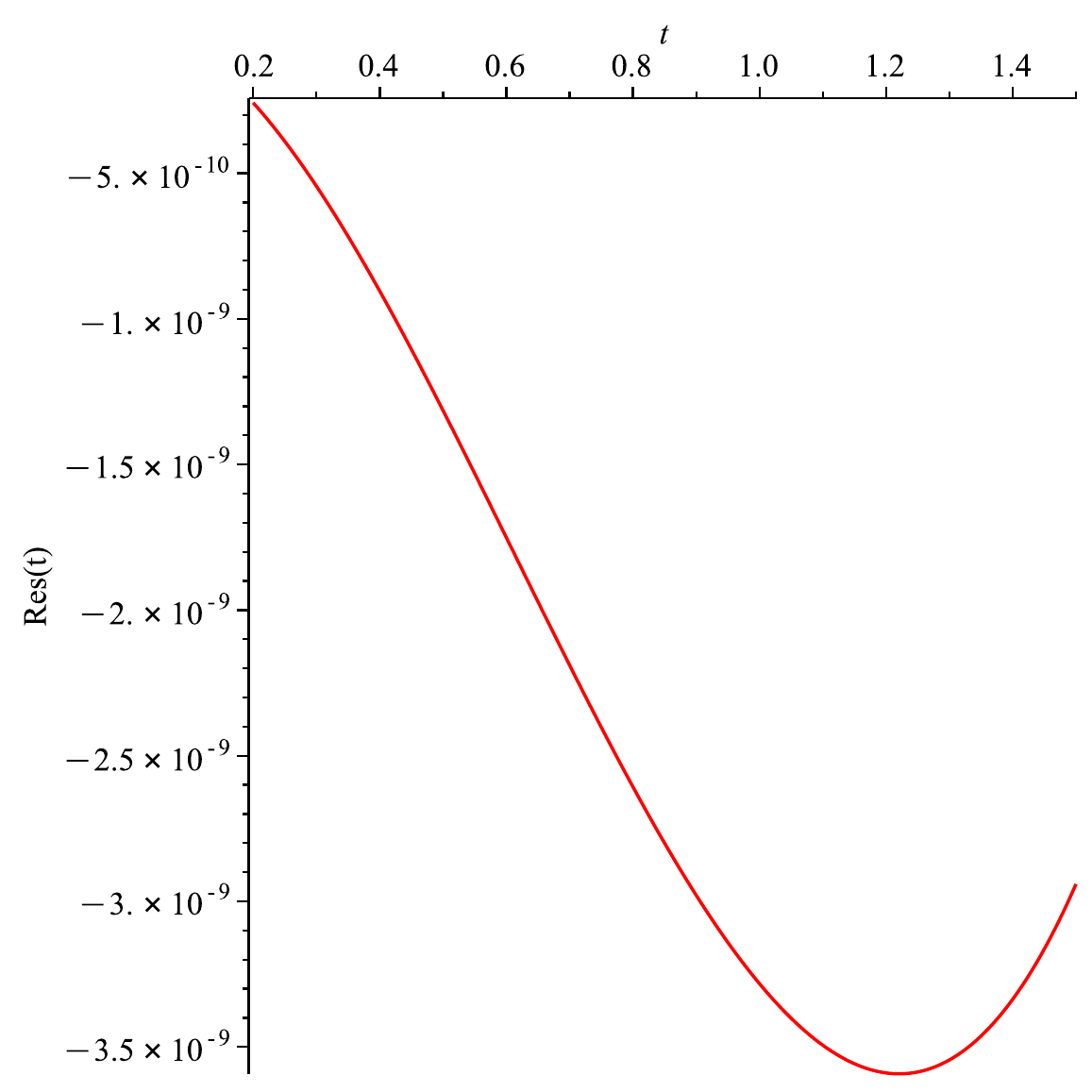}
    \caption{Residual with respect to the exact solution}
    \end{subfigure}
    \caption{Simulation results of Example \ref{ex:1}}
    \label{fig:ex1}
\end{figure}

\end{example}

\begin{example}
\label{ex:2}
For the second example, we consider the following DOFDE \cite{mashayekhi2016numerical}:
\begin{equation*}
\int_{0}^{1} 6\theta (1-\theta) \, D^\theta \, u(t) \, d\theta + \frac{1}{10}u(t) = 0,
\end{equation*}
with the initial condition \( u(0) = 1 \). This problem does not have an exact solution in the time-domain space. Therefore, after simulating this problem with \( \gamma = 10^{12} \) and $t\in[0,1]$, we report the simulated results for different numbers of basis functions in Table \ref{tbl:ex2}. The residual function of the obtained solution, along with the learned approximation with \( d = N = 20 \), is shown in Figure \ref{fig:ex2}. The simulated results shows a good agreement with previous works \cite{mashayekhi2016numerical}
\begin{figure}[ht]
    \centering
    \begin{subfigure}[b]{0.45\textwidth}
        \centering
        \includegraphics[width=\textwidth]{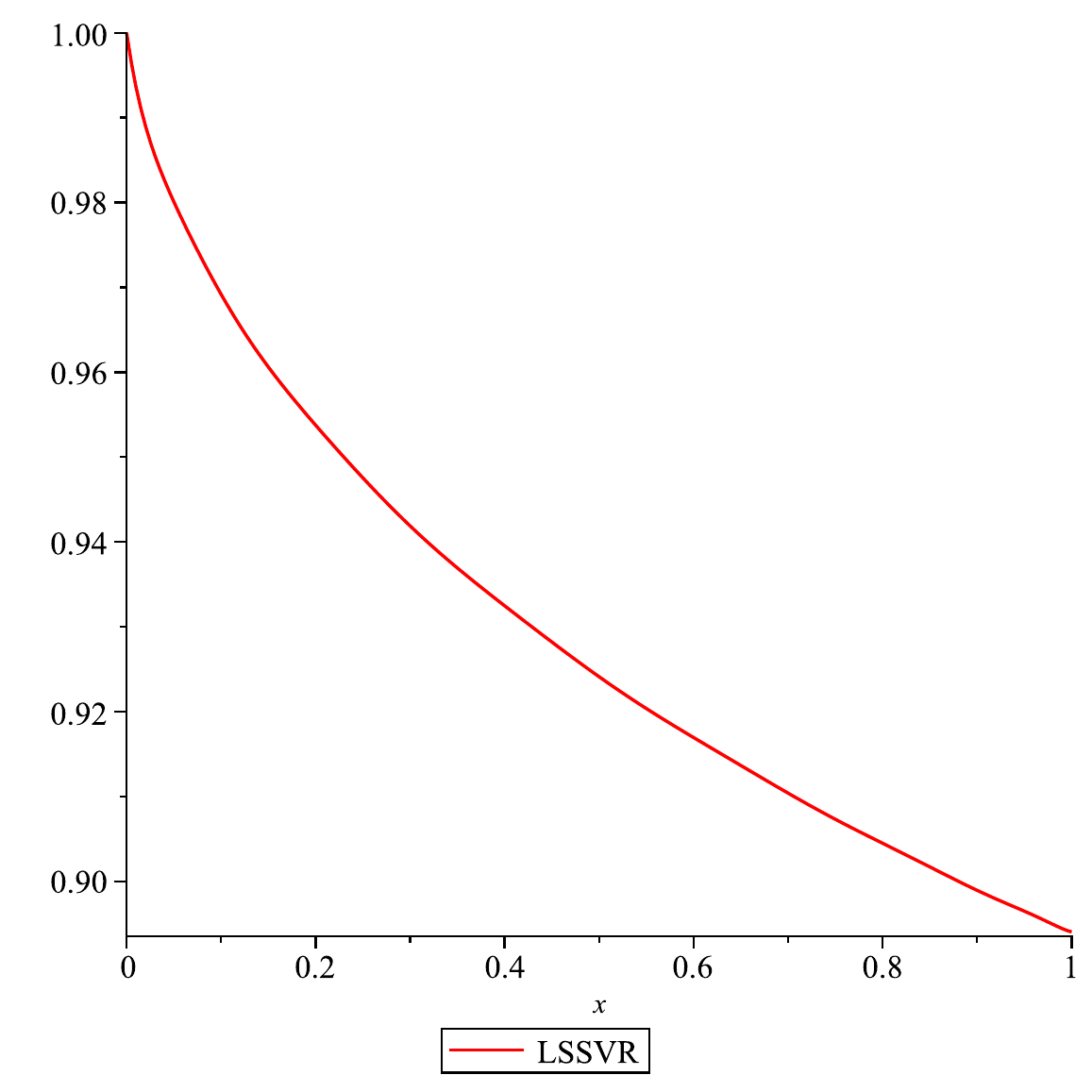}
    \caption{Predicted solution}
    \end{subfigure}
    \hfill
    \begin{subfigure}[b]{0.45\textwidth}
        \centering
        \includegraphics[width=\textwidth]{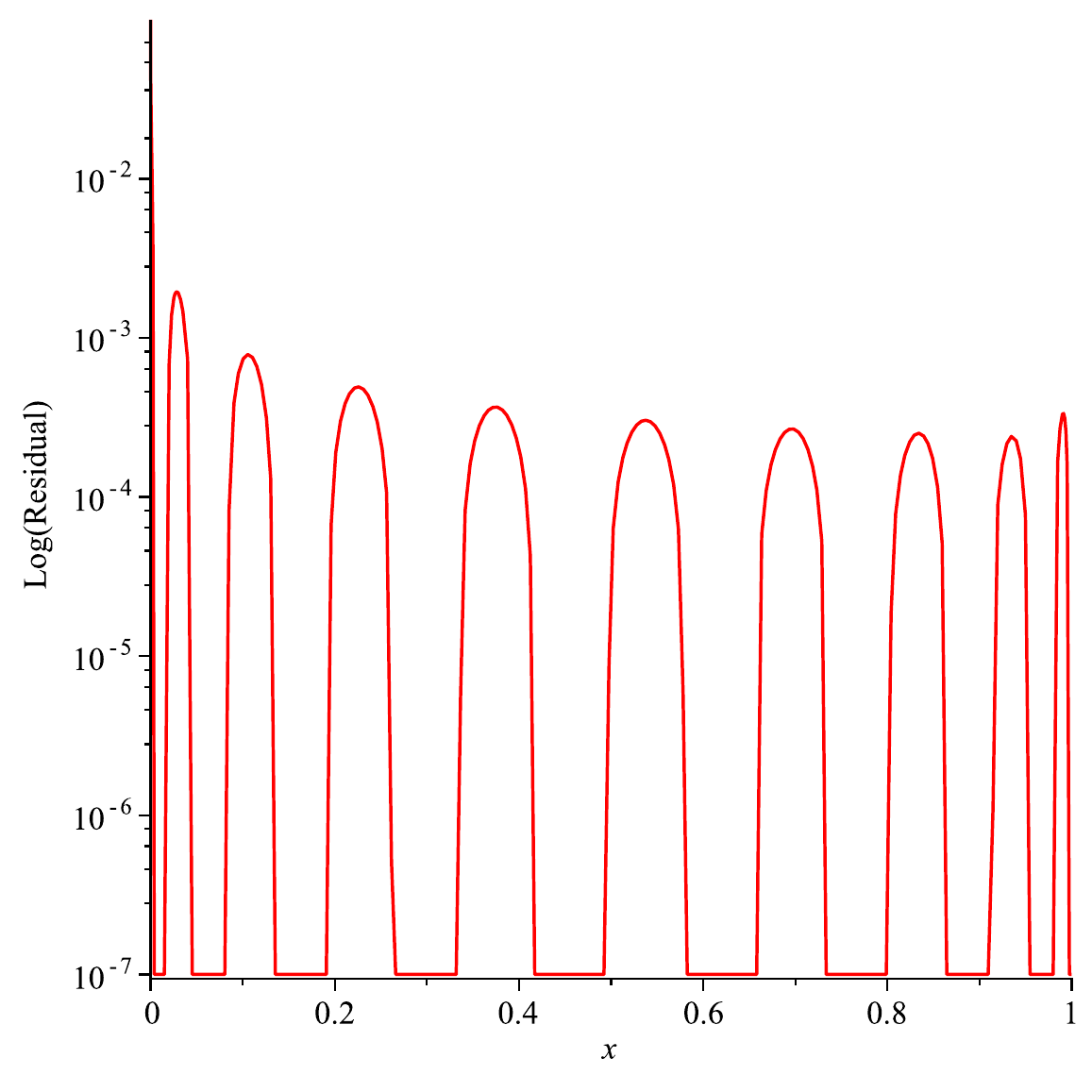}
    \caption{Residual function in logarithmic scale}
    \end{subfigure}
    \caption{Simulation results of Example \ref{ex:2}}
    \label{fig:ex2}
\end{figure}

\begin{table}[ht]
\centering
\begin{tabularx}{\textwidth}{@{}XXXX@{}}
\toprule
$t$ & $d=10$ & $d=15$ & $d=20$ \\ \midrule
0.1 & 0.969161897444615 & 0.969185065765532 & 0.969137509011574 \\
0.2 & 0.953959557214783 & 0.953761401002633 & 0.953600844737208 \\
0.3 & 0.941670493869908 & 0.941886826855227 & 0.941941233621975 \\
0.4 & 0.932544469093204 & 0.932475004172846 & 0.932422818680849 \\
0.5 & 0.924441458486824 & 0.924113321766607 & 0.924122395416036 \\
0.6 & 0.916738219364862 & 0.916943568203690 & 0.916912664051348 \\
0.7 & 0.910362673712272 & 0.910397667462537 & 0.910418004162028 \\
0.8 & 0.904721558913845 & 0.904511874231876 & 0.904457638628798 \\
0.9 & 0.898916530968376 & 0.899008078487048 & 0.899027710512671 \\
1.0 & 0.893997130689097 & 0.894032641945487 & 0.894007351849340 \\ \bottomrule
\end{tabularx}%
\caption{The predicted solutions after simulating Example \ref{ex:2} using the proposed method for different values of $d$.}
\label{tbl:ex2}
\end{table}

\end{example}
\subsection{Partial DOFDEs}
In this section, we consider DOFDEs with unknown solutions that depend on two independent variables. For all of the problems in this section, we consider Gaussian quadrature discretization $Q=7$.
\begin{example}
\label{ex:3}
Consider a distributed fractional partial differential equation in the form:
\begin{equation*}
\int_{0}^{1} \Gamma(3-\theta) \frac{\partial^{\theta} u}{\partial t^{\theta}}(x, t) \, d\theta = \frac{\partial^{2} u}{\partial x^{2}}(x, t) + 2 t^{2} + \frac{2 t x (t-1)(2-x)}{\ln t},
\end{equation*}
with initial and boundary conditions:
\begin{align*}
u(x, 0) &= 0, \quad &0 < x < 2,\\
u(0, t) &= u(2, t) = 0, \quad &0 < t \leq 1.
\end{align*}
This problem has the analytical solution \( u(x, t) = t^{2} x (2-x) \) \cite{moghaddam2019numerical,morgado2015numerical}. We simulate this problem using \( d_x = d_t = 3 \) basis functions and 9 training points in the problem domain, which are obtained from the roots of the basis functions. The simulation result for this problem is depicted in Figure \ref{fig:ex3}.

\begin{figure}[ht]
    \centering
    \begin{subfigure}[b]{0.45\textwidth}
        \centering        \includegraphics[width=\textwidth]{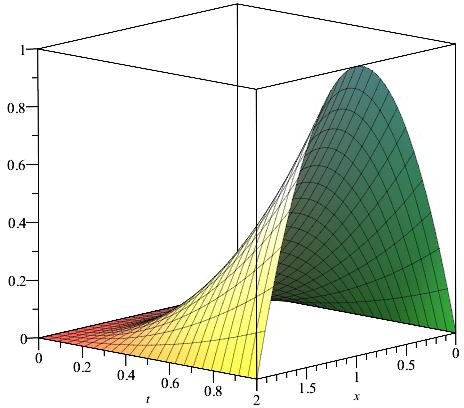}
\caption{Predicted solution}
    \end{subfigure}
    \hfill
    \begin{subfigure}[b]{0.45\textwidth}
        \centering
        \includegraphics[width=\textwidth]{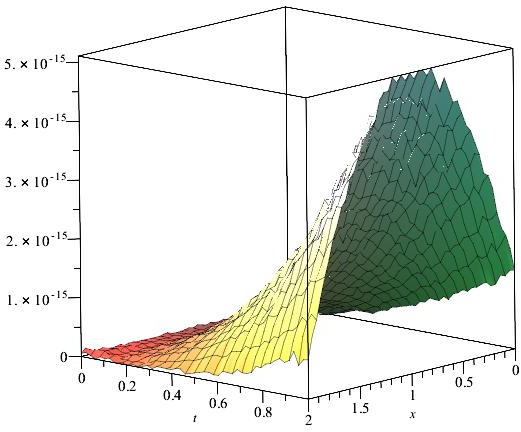}
    \caption{Residual with respect to the exact solution}
    \end{subfigure}
    \caption{Simulation results of Example \ref{ex:3}}
    \label{fig:ex3}
\end{figure}

\end{example}

\begin{example}
\label{ex:4}
For the final experiment, we consider the following partial DOFDE \cite{morgado2015numerical}:
\begin{equation*}
\int_{0}^{1} \Gamma(3.5-\theta) \frac{\partial^{\theta} u}{\partial t^{\theta}}(x, t) \, d\theta = \frac{\partial^{2} u}{\partial x^{2}}(x, t) + u(x,t)^2+\frac{15 \sqrt{\pi} \left(t-1\right) t^{\frac{3}{2}}}{8\ln\left(t\right)} x \left(x-1\right)-2t^{\frac{5}{2}}-t^{5}x^{2} \left(x-1\right)^{2},
\end{equation*}
with the exact solution \(u(x,t) = t^2 \sqrt{t} \, x (x - 1)\), which yields the initial and boundary conditions:
\begin{equation*}
\begin{aligned}
    u(x, 0) &= 0, \quad 0 < x < 1,\\
    u(0, t) &= u(1, t) = 0, \quad 0 < t \leq 1.
\end{aligned}
\end{equation*}
Using the proposed approach to solve this problem, we employed \(d_x=3\), \(d_t=15\) with \(N=45\) training points. The approximated solution is depicted in Figure \ref{fig:ex4}. Table \ref{tbl:ex4} also reports the approximated solution at specific points in the problem domain.

\begin{figure}[ht]
    \centering
    \begin{subfigure}[b]{0.45\textwidth}
        \centering
        \includegraphics[width=\textwidth]{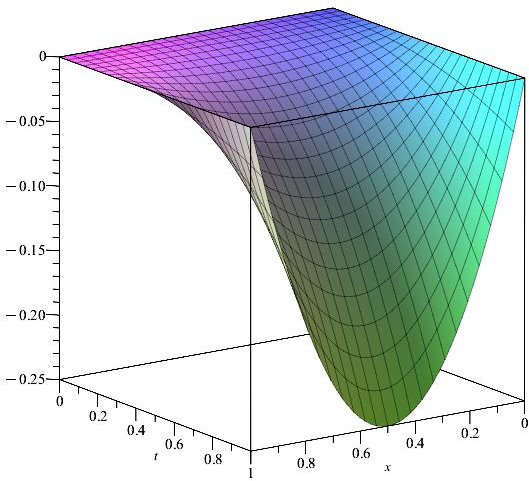}
        \caption{Predicted solution}
    \end{subfigure}
    \hfill
    \begin{subfigure}[b]{0.45\textwidth}
        \centering
        \includegraphics[width=\textwidth]{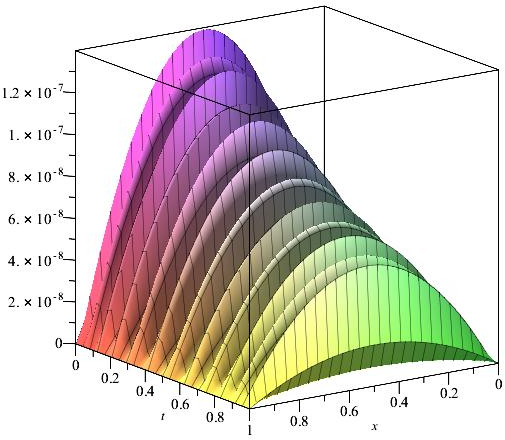}
        \caption{Residual with respect to the exact solution}
    \end{subfigure}
    \caption{Simulation results of Example \ref{ex:4}}
    \label{fig:ex4}
\end{figure}

\begin{table}[ht]
\centering
\begin{tabular}{@{}rcccccc@{}}
\toprule
$x \backslash t$ & 0.0 & 0.2 & 0.4 & 0.6 & 0.8 & 1.0 \\ \midrule
0.0 & 0.0 & 0.0 & 0.0 & 0.0 & 0.0 & 0.0 \\
0.1 & 0.0 & -0.00160994 & -0.00910733 & -0.02509692 & -0.05151899 & -0.08999999 \\
0.2 & 0.0 & -0.00286211 & -0.01619081 & -0.04461675 & -0.09158932 & -0.15999999 \\
0.3 & 0.0 & -0.00375652 & -0.02125044 & -0.05855948 & -0.12021099 & -0.20999998 \\
0.4 & 0.0 & -0.00429316 & -0.02428622 & -0.06692512 & -0.13738398 & -0.23999998 \\
0.5 & 0.0 & -0.00447204 & -0.02529814 & -0.06971366 & -0.14310832 & -0.24999998 \\
0.6 & 0.0 & -0.00429316 & -0.02428622 & -0.06692512 & -0.13738398 & -0.23999998 \\
0.7 & 0.0 & -0.00375652 & -0.02125044 & -0.05855948 & -0.12021099 & -0.20999998 \\
0.8 & 0.0 & -0.00286211 & -0.01619081 & -0.04461675 & -0.09158932 & -0.15999999 \\
0.9 & 0.0 & -0.00160994 & -0.00910733 & -0.02509692 & -0.05151899 & -0.08999999 \\
1.0 & 0.0 & 0.0 & 0.0 & 0.0 & 0.0 & 0.0\\
\bottomrule
\end{tabular}%
\caption{The predicted value for some test points in the problem domain for Example \ref{ex:4}.}
\label{tbl:ex4}
\end{table}
\end{example}

\section{Conclusion}
In this study, we developed a physics-informed machine learning approach for numerically solving distributed-order fractional differential equations. We specifically tailored the Least Squares Support Vector Regression algorithm to capture the intricate dynamics of these problems. By employing Gegenbauer polynomials as the kernel function, the LSSVR was optimized to deliver precise predictions of the unknown solutions. We also demonstrated the effectiveness of the Gaussian quadrature for approximating the integral components of the equations and leveraged the properties of the Caputo derivative to enhance computational efficiency.

Our numerical experiments included two ordinary and two partial DOFDEs, where the proposed framework successfully approximated their solutions. For problems with known exact solutions, our method showed high accuracy, while for problems without exact solutions, we provided the simulated results obtained from LSSVR which are in good agreement with previous works. Future research could explore the use of alternative kernel functions or generalize the approach with fractional basis functions. Additionally, integrating advanced hyperparameter optimization techniques to fine-tune model parameters presents a promising direction for further study.





\begingroup
\setstretch{1.0}
\bibliographystyle{customstyle}
\bibliography{references}
\endgroup


\end{document}